# A Note on Geometric Calibration of Multiple Cameras and Projectors

Tomislav Petković*, Simone Gasparini**, and Tomislav Pribanić*
* University of Zagreb, Faculty of Electrical Engineering and Computing, Unska 3, Zagreb, Croatia
** University of Toulouse, Toulouse INP – IRIT, Toulouse, France
tomislav.petkovic.jr@fer.hr

**Geometric calibration of cameras and projectors is an essential step that must be performed before any imaging system can be used. There are many well-known geometric calibration methods for calibrating systems comprised of multiple cameras, but simultaneous geometric calibration of multiple projectors and cameras has received less attention. This leaves unresolved several practical issues which must be considered to achieve the simplicity of use required for real world applications. In this work we discuss several important components of a real-world geometric calibration procedure used in our laboratory to calibrate surface imaging systems comprised of many projectors and cameras. We specifically discuss the design of the calibration object and the image processing pipeline used to analyze it in the acquired images. We also provide quantitative calibration results in the form of reprojection errors and compare them to the classic approaches such as Zhang's calibration method.**

*Keywords; geometric calibration; camera calibration; projector calibration; system of multiple cameras and projectors; reprojection error*

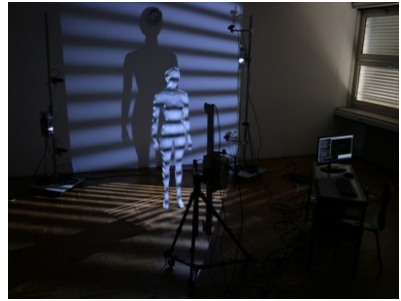

Figure 1. A 3D surface scanning system comprised of three projectors and six cameras

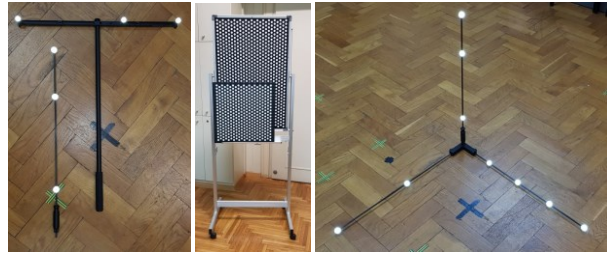

Figure 2. Various calibration objects: calibration wands (left), calibration boards (center), and a 3-axis spatial calibration object (right)

## I. Introduction

Both geometric calibration and photometric calibration of cameras and projectors have many applications including remote sensing [1], photogrammetry [2,3], 3D surface imaging and profilometry [4,5], motion capture [6,7,8], tiled displays [9,10,11], and many others. The task of geometric calibration [12,13,14] is to determine intrinsic and extrinsic parameters [12,13] of cameras and projectors, while the task of photometric calibration [15] is to establish an exact relation between measured values and physical units. In geometric calibration intrinsic parameters describe the internal properties of a camera or a projector. Extrinsic parameters describe their spatial position w.r.t. some reference coordinate system, usually in terms of a 3D rotation matrix and a translation vector. Geometric calibration is therefore an essential step which must be performed before any imaging system can be used as a measurement tool. It should be as simple as possible while retaining required accuracy in estimating imaging system parameters, especially when a system of multiple cameras and projectors is considered (Fig. 1).

The usual approach to geometric calibration is to record a calibration object such as the ones shown in Fig. 2 [12,13]. Relevant parameters of the imaging system are then estimated from the data by minimizing the reprojection error between the acquired images and the expected appearance of the calibration object obtained through an imaging model.

Differences between various proposed calibration methods stem from the fact that many different calibration objects are proposed. One amongst many possible classifications [12] is based on the degrees of freedom of the calibration object: (a) one dimensional objects such as calibration wands (Fig. 2 left) are most often used for simultaneous calibration of multiple cameras [6,8,16,17]; (b) two dimensional objects such as planar calibration boards (Fig. 2 center) are currently considered the state of the art in calibrating single cameras and stereo pairs [14,18,19,20]; and (c) three dimensional calibration objects such as calibration cages are mostly limited to laboratories for precise calibrations [8].

The most significant limitation of existing calibration techniques is that the calibration object must be completely visible for it to be considered as a valid input; examples include standard implementations available in OpenCV [19] or in Matlab [20] where the chessboard pattern is rejected if partially visible. This presents a significant limitation for the calibration of a system comprised of many cameras and projectors (Fig. 1) as it is difficult or even impossible to position the calibration object so that it is entirely visible in all the views.

In this work we describe a calibration procedure routinely used in our laboratory [22] to calibrate an imaging system comprised of many cameras and projectors. We have introduced a first implementation of this method in [23] on which we here expand by describing the used image processing chain and by providing quantitative results. The proposed method uses a double-sided planar calibration board. A calibration pattern on both sides is comprised of white circles on dark background arranged in a hexagonal lattice and includes additional marking to easily identify which side of the board is observed [23]. The proposed image processing pipeline is able to extract reliable calibration points even if the calibration board is partially visible which presents a significant improvement compared to standard calibration procedures which require that the whole calibration object and which discard partial views.

This work is organized as follows: in Section II. we give a detailed overview of existing camera and projector calibration methods. The proposed calibration method is presented in section III. while results and discussion are presented in section IV. We conclude in Section V.

## II. BACKGROUND

When considering geometric calibration of multiple cameras and projectors one must note that a projector may be modelled as an inverse camera [24] and that existing cameras in the imaging system may be used to acquire images as seen by the projector [24,25]. Therefore, we will first give a brief overview of geometric calibration methods for cameras followed by methods for projectors.

### A. Camera Calibration

A most commonly used imaging model is the pinhole camera model [12] which is almost always augmented by including lens distortion [12,14,19], although sometimes other imaging models such as fisheye camera model [26,27] may be used. Regardless of the model which is used all camera calibration procedures share the same steps: (a) acquire many images of a calibration object; (b) process acquired images to extract the required data such as image coordinates of the calibration object or its characteristic points; and (c) minimize the reprojection error and find the optimal parameters of the imaging model through optimization. The first two steps determine the ease of use in real-world scenarios. In practice, most popular approaches include wand calibration [6,8,16,17] and calibration using a planar calibration board [14,28,29].

In wand calibration we use a wand containing several spherical markers made of reflective material (Fig. 2). An important property of spherical markers is that their center can be easily extracted by detecting circle centers in 2D images of each camera separately. This enables fast acquisition of calibration data simply by waving a calibration wand in a form of wand dance [8].

In planar calibration we use a planar calibration board containing a known calibration pattern (Fig. 2). A commonly used pattern is the checkerboard [19,20] from which corner points can be easily detected. Other patterns such as circles arranged in a regular lattice are also used [8,25]. All planar patterns can be easily printed on plain paper which makes them a method of choice for calibrating single cameras and stereo pairs. The main problem when using a planar calibration object for calibrating a system of multiple cameras is that it is visible from one side only, unlike a calibration wand which is visible from all sides simultaneously. This limitation may be improved upon by using a double-sided calibration board [23] where a rigid transformation between patterns on each side is known. This, however, comes at the cost of more complex and expensive manufacturing which may require an external contractor.

For completeness here we also mention auto- or self-calibration [13,31] which does not require a calibration object and which is often used in photogrammetry or in virtual environments.

### B. Projector Calibration

A major obstacle in projector calibration is that the projector cannot acquire images. Therefore the cameras of imaging system must be used to acquire the required data instead [24,25]. This is usually achieved by having a projector project some coded pattern which enables a mapping of camera images into the coordinate system of projector's image.

First approaches to projector calibration assumed a calibrated camera is used [32,33,34,35,36]. Then some code is projected onto planar surfaces either to directly find a homography or to indirectly get the required data for calibration: fiducial markers are used in [32,33,36], chessboard pattern in [34], and a specially designed pattern in [35]. Such approaches require twice the effort and do not scale well when calibrating systems with many cameras and projectors.

Modern approaches [24,25,37,38,23] follow the proposal of Zhang and Huang [24] and employ ideas from structured light scanning [4]. A projector should project a code which directly embeds projector's row and column coordinates. Decoding the projected code then provides a direct mapping between camera's and projector's image coordinate systems. A significant advantage of directly measuring the mapping is that camera does not need to be calibrated in advance. Differences in listed approaches stem from the used code: in [24,25] a standard multiple phase shifting is used, [37] uses Gray code, and [38] a combination of them. Note that phase shifting is a better choice as it inherently provides subpixel precise mappings while Gray code requires interpolation to achieve subpixel precision. Another significant difference is the choice of the planar calibration pattern: some works use a standard checkerboard [24,37] while others use a regular lattice of bright circles on dark background [25,23] which makes the projected code easier to decode.

Other less common approaches to projector calibration include self-calibration [39] where multiple cameras are used to calibrate a single projector, and the use of a semi-transparent sheet [40] to make the code projected by the projector simultaneously visible by multiple cameras in an imaging system.

III. CALIBRATION PROCEDURE

We propose a calibration procedure for multi-projector multi-camera 3D scanner which is based on the following principles: (1) a calibration object must be usable even if it is only partially visible; (2) calibration points on the calibration object should be automatically detected and annotated; and (3) cameras will provide a means for projector to indirectly capture an image of the calibration board using a method proposed by Zhang and Huang [24]. We use a planar calibration object so that Zhang's planar method [14] can be used to obtain a quick initial calibration for each camera and projector. However, Zhang's method provides decoupled calibration parameters therefore the final calibration must include some form of simultaneous minimization of the reprojection errors for all cameras and projectors using bundle adjustment [21].

### A. Calibration Pattern

We propose to use a calibration pattern comprised of bright circles on a dark background arranged in a regular hexagonal lattice (Fig. 3). The pattern may be used on a single-sided calibration board, but we chose a double-sided calibration board to speed up calibration and to reduce the number of acquisitions. Such board may be professionally manufactured, or it may be self-constructed by printing out the pattern and carefully gluing it to both sides. If printed and glued some additional method of measuring the transformation between sides must be used; for the board shown in Fig. 1 we have used MicroScribe G2LX [30] digitizing system to find the transformation.

To make the calibration simpler we also add marks in a form of dark rings which are used to identify the center and the side of the calibration board. To allow for any spatial orientation of the calibration board the markings are limited to an elemental hexagon and are selected in a way which makes all possible planar rotations of the board unique. Consider a hexagonal cell where the central marker `a` is surrounded by six neighbors which are arranged clockwise starting from `b` (Fig. 4). If the orientation of the calibration board is arbitrary and we do not know it then when identifying the cell any of the surrounding elements `bcdefg` may take the starting role of `b`. Therefore, the codes for identifying sides must selected so all rotational permutations of the surrounding six elements `bcdefg` are mutually different thus making all four used codes uniquely decodable.

### B. Recording the Calibration Pattern

The calibration board should be recorded in many positions with respect to the cameras; a minimum is two different positions for each camera and projector [14]. Then for each position of the calibration board we have to acquire as many images as selected structured light code

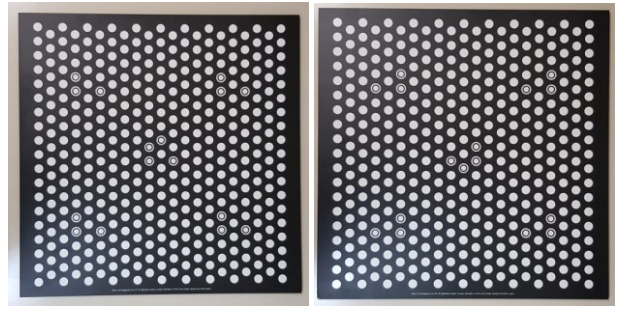

Figure 3. Two sides of a double sided planar calibration board (front and back). Note the markers in forms of inscribed black rings which identify the board side and the central point.

```
            1        0
      1     0   0    0    1
         b      0    1
      g     c   1         1
         a      0    0
      f     d   1    0    1
         e      0    0
            1    1
            0    0
```

Figure 4. One hexagonal cell and corresponding values for markings denoting the central points on front and back side and sided. 1 indicates the ring is present and 0 it is not. Note that all planar rotations of four used codes are mutally distinguishable.

requires. We propose to use our multiple phase shifting patterns which enable simultaneous acquisition in a system comprised of an arbitrary number of projectors [41]. Then for each position of the calibration board from the acquired images we compute images as seen by each projector containing illuminated parts of the board. Furthermore, if more than one camera observed the illuminated part of the board for some projector then the number of data points for that particular projector is increased, i.e. we get multiple measurements from one complex acquisition. Note that the method presented in [41] does not require acquisition of separate camera image when all projectors are turned off; this image is computed instead.

### C. Pattern Processing

Once the images are acquired and the structured light code is decoded using the approach from [41,42] we have to identify the circle centers and to construct a hexagonal lattice of calibration points. To make this fully automated we propose to use multi-scale analysis where on each scale the eigenvalues of the Hessian matrix are used to construct a *circleness* measure in a similar way as was proposed by Frangi et al. [43] for ridge detection by using a *vesselness* measure.

Let $I(x,y)$ denote the input image and let $G_\sigma(x,y)$ be a Gaussian kernel at scale σ. The Hessian matrix for each pixel $(x,y)$ is then

$$H_\sigma(x,y) = \begin{bmatrix} I * \frac{\partial^2 G_\sigma}{\partial x^2} & I * \frac{\partial^2 G_\sigma}{\partial x \partial y} \\ I * \frac{\partial^2 G_\sigma}{\partial y \partial x} & I * \frac{\partial^2 G_\sigma}{\partial y^2} \end{bmatrix}, \quad (1)$$

where $(x,y)$ dependencies are omitted from $I$ and $G_\sigma$. For the Hessian matrix given by Eq. (1) we compute the eigenvalues $\lambda_{1,2}$ for each pixel $(x,y)$ and order them so that

$$|\lambda_1| \geq |\lambda_2|. \quad (2)$$

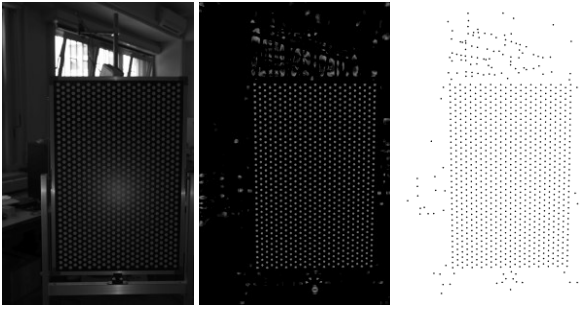

Figure 5. Detection of circles using circleness measure: input image (left), circleness at all scales (center), detected centers (right).

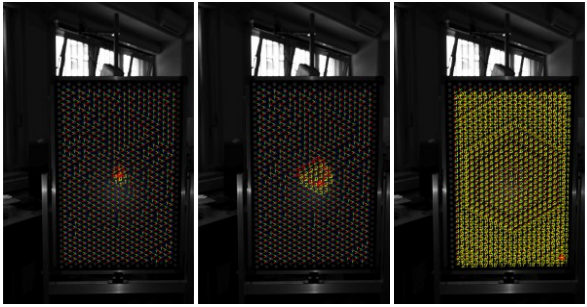

Figure 6. Grid construction: first cell (left), grid after 27 cells are processed (center), and final grid (right).

The circleness measure to detect bright circles on dark background is then

$$C_\sigma(x,y) = \left(1 - \exp\left(-\alpha \frac{\lambda_2^2}{\lambda_1^2}\right)\right)\left(1 - \exp\left(-\beta(\lambda_1^2 + \lambda_2^2)\right)\right), \quad (3)$$

for $\lambda_{1,2} < 0$ and is zero otherwise. In Eq. (3) we have fixed parameter $\alpha$ to $\frac{1}{4\log 2}$ and $\beta$ is determined using the ridge model from [44] under the assumption of an ideal circle instead of an ideal ridge (see Eqs. (3) and (14) in [44]). Note that other choices of parameters are possible as explained in [43,44]. The circleness measure of Eq. (3) is by its definition limited to the [0,1] interval and is easily integrated over all scales using

$$C(x,y) = \max_{\text{over all scales } \sigma} C_\sigma(x,y), \quad (4)$$

In Eq. (4) scales are selected to encompass all expected circle sizes. The final result of this analysis is an image (Fig. 5 center) where circle's centers are local maxima. Note that non-maximum suppression must be used to suppress multiple detections, especially if the image exhibits strong perspective distortion: from the scale-space analysis radii for all detections are known so we delete all weaker circle centers having smaller vesselness value and falling within the radius of the locally strongest one (Fig. 5 right).

Once all centers are extracted a hexagonal grid must be constructed. We propose to do this in a two-pass procedure. In the first pass we construct all viable local elementary hexagons (see Fig. 4). In the second pass all constructed hexagons are assembled into a complete grid. This two-pass approach allows us to construct a usable grid even if the calibration board is only partially visible.

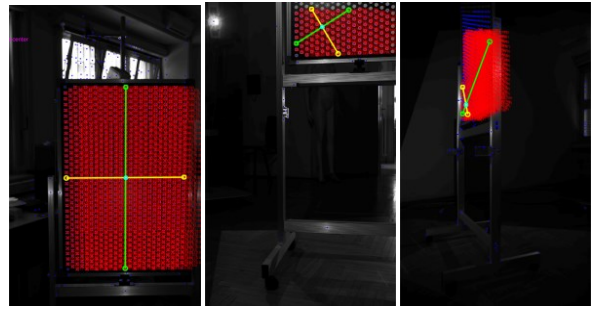

Figure 7. Constructed calibration grids: regular grid (left), and partial grids where markers are not visible or cannot be decoded (center and right) so coordinate systems are arbitrary oriented. Note the extreme perspective distortion in the rightmost example.

In the first pass a local hexagon is constructed around each detected circle center as follows: (1) find opposing pairs for the six closest neighboring circles; (2) for each opposing pairs compute the normalized distance to the central point; (3) select three pairs which have the shortest normalized distance. The key point of this pass is distance normalization: for each opposing pair the line connecting them should pass through the center of the central circle and should intersect the edge of that circle thus forming a chord, more specifically a diameter. The length of this chord changes depending on the amount of perspective distortion. If we divide the length between opposing points by the chord length, then distances are normalized and can be compared to sort the points. Without this normalization the step (3) could produce a deformed local lattice where we select an opposing pair which is closest in the image space, but not in the real space. Once the hexagon cell is constructed the embedded code is also decoded using a simple heuristic inscribed ring detection procedure so we can identify if a cell is a central point or if a cell is located on front or back side. Note that the first pass also acts as a natural filter: all found cells which cannot be matched to a hexagon are discarded.

In the second pass we start from the best local hexagonal cell. In order of desirability the starting cell is one containing the board's central point, then a cell indicated by repeated side marks, and finally any cell if no marks are visible (or if marks conflict). We then push all unprocessed neighboring cells whose centers are at **bcdefg** points (Fig. 4). into a FIFO queue. The constructed grid is expanded when a cell is popped from the queue, but only if the cell can be properly fitted with previously placed cells. This fitting includes planar rotation of each cell; for an example see Fig. 6 left and center where it is visible that local coordinate systems are not initially aligned, but they become aligned after all cells are processed. This behavior is expected as the basic hexagonal element of the proposed grid is highly symmetrical making it impossible to determine the global orientation from local data (see Fig. 7).

Once both steps are complete, we have obtained corresponding coordinate pairs comprised of ideal planar coordinates on the calibration grid and of camera or projector coordinates in subpixel accuracy. This pairs can then immediately be used in Zhang's calibration method [14] to obtain an initial calibration, however note that Zhang's calibration method [14] is intended for calibration

TABLE I. REPROJECTION ERRORS FOR A SIX CAMERAS THREE PROJECTORS IMAGING SYSTEM. ALL VALUES ARE IN PX.

| ID | Zhang's method | Bundle adjustment |
|---|---|---|
| CAM1 | 0.21 ± 0.30 | 2.87 ± 2.28 |
| CAM2 | 0.21 ± 0.15 | 4.35 ± 2.25 |
| CAM3 | 0.23 ± 0.25 | 1.18 ± 0.80 |
| CAM4 | 0.23 ± 0.16 | 1.81 ± 1.08 |
| CAM5 | 0.20 ± 0.16 | 2.74 ± 1.96 |
| CAM6 | 0.21 ± 0.25 | 2.36 ± 1.53 |
| PRJ1 | 0.22 ± 0.15 | 1.44 ± 0.97 |
| PRJ2 | 0.22 ± 0.15 | 0.64 ± 0.50 |
| PRJ3 | 0.26 ± 0.36 | 1.64 ± 1.27 |

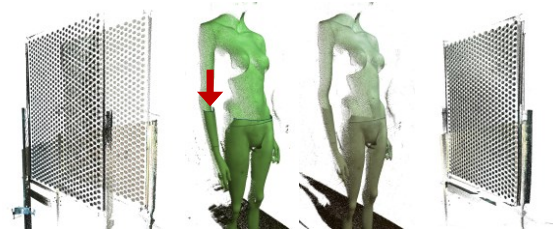

Figure 8. Point-cloud reconstructions: On the left are reconstructions using calibration data of Zhang's method when a single position of the board was selected as referenced. On the right are same reconstructions using calibration data from bundle adjustment.

of a single camera and it cannot provide a reliable calibration for a multi-view system.

*D. Calibration*

After a pattern is processed all found corresponding point pairs are not necessarily expressed in the same coordinate system: (a) if the central marker is not visible then coordinate axes may be arbitrarily set; and (b) if a side markers are not visible then differentiating between front and back side is impossible. To solve these issues, we propose to use projector data to find correspondences which then enable computation of 2D rigid registrations to align found points to the ideal grid. In other words, extracted projector coordinates are used as point descriptors and all data from partially visible boards is automatically registered to valid data for which both the center location and the side of the board are known. Sometimes even after this procedure there are leftover unaligned partial views: we leave the choice to the user to either discard them or to manually realign the grid coordinates. The data obtained in this way is fed into the bundle adjustment procedure to minimize the reprojection error

$$\mathbf{e} = \mathbf{x}_{px} - \mathbf{x}(X_{3D}, K_c, R_c, t_c, S_k, u_k), \quad (5)$$

where $K_c$ are intrinsic parameters (camera matrix and distortion parameters), $R_c$ and $t_c$ are extrinsic parameters (rotation and translation) of each camera and projector, and $S_k$ and $u_k$ (rotation and translation) are positions of the calibration board. If the system has $N$ cameras and $M$ projectors and if $K$ positions of the calibration board are recorded then there are $6(N+M)$ intrinsic parameters $K_c$, while the total number of extrinsic parameters $R_c$, $t_c$ and $S_k$, $u_k$ is reduced to $6(N+M+K-1)$ one as one position must be taken as a reference.

## IV. RESULTS AND DISCUSSION

The presented calibration procedure is used in our laboratory to routinely calibrate systems of up to six cameras and three projectors such as one shown in Fig. 1. Reprojection errors for one calibration of such system are shown in Table I: errors are in the order of pixels, which is sufficient for everyday use. In this example three positions of a double-sided calibration board were used which gives three planar grids for calibration, each containing 838 circles if fully visible. Note that if more than one camera views the board then the number of points for a particular projector is increased as there are multiple simultaneous measurements. If a more precise calibration is required more positions should be recorded. Note that calibration errors are smaller for Zhang's calibration method. This is due to the fact that Zhang's method allows more degrees of freedom than bundle adjustment so model is easier to fit to the data, i.e. in Zhang's method each grid position has its own rotation and translation parameters while in bundle adjustment these are shared. This is important in practice as only complete calibration allows us to perform reconstructions, e.g. in Fig. 8 when recording a double sided calibration board using opposing cameras reconstruction is impossible without full bundle adjustment. Even when a common a trick is used to express all extrinsic camera parameters using one selected position of the calibration board point clouds obtained from different viewpoints are not perfectly aligned as indicated by the red arrow in Fig. 8. In other words, a better reprojection error for Zhang's method is paid for with the uncertainty in the extrinsic parameters.

## V. CONCLUSION

We have presented a practical approach to calibrating a system comprised of multiple cameras and projectors. Presented approach is easy to use in practice and allows us to calibrate a complex imaging system quickly and efficiently.

ACKNOWLEDGMENT

This work has been supported by the Croatian Science Foundation under the grant number HRZZ-IP-2019-04-9157 (3D-CODING).